\documentclass[letterpaper, 10 pt, conference]{ieeeconf}  

\IEEEoverridecommandlockouts                              

\usepackage{graphicx}      
\usepackage{comment}
\usepackage[utf8]{inputenc} 
\usepackage{tikz}
\usepackage{algorithmic}
\usepackage{graphicx}
\usepackage{subfig}
\usepackage{amsfonts}
\usepackage{textcomp}
\usepackage{xcolor}
\usepackage{cite}
\usepackage{csquotes}
\usepackage{multicol}
\usepackage{hyperref}
\usepackage{xcolor}
\usepackage{siunitx}
\usepackage{hyperref} 
\hypersetup{backref=true,       
    pagebackref=true,               
    hyperindex=true,                
    colorlinks=true,                
    breaklinks=true,                
    urlcolor= black,                
    linkcolor= blue,                
    bookmarks=true,                 
    bookmarksopen=false,
    filecolor=black,
    citecolor=blue,
    linkbordercolor=blue
}

\usepackage{mathtools}

\usepackage{caption}
\usepackage{balance}
\usepackage{soul} 

\newcommand\numeq[1]%
  {\stackrel{\scriptscriptstyle(\mkern-1.5mu#1\mkern-1.5mu)}{=}}

\newtheorem{remark}{Remark}[section]

\newtheorem{assumption}{Assumption}
\newtheorem{problem}{Problem}
\usepackage{hyperref}

\title{\LARGE \bf
Online Non-linear Centroidal MPC for Humanoid Robots Payload Carrying with Contact-Stable Force Parametrization
}

\author{Mohamed Elobaid$^1$, Giulio Romualdi$^1$, Gabriele Nava$^1$, Lorenzo Rapetti$^{1,3}$, \\ Hosameldin Awadalla Omer Mohamed$^{1,2}$ and Daniele Pucci$^{1,3}$
\thanks{$^{1}$ Artificial and Mechanical Intelligence \emph{AMI} (Italian Insititute of Technology); Genoa, Italy {\tt\small {\{firstname.lastname\}@iit.it}}.}%
\thanks{$^{2}$ Mechanical engineering department; Politecnico di Milano,  Milano, Italy {\tt\small {\{firstname.lastname\}@polimi.it.}}}
\thanks{$^{3}$ Machine Learning and Optimisation, The University of Manchester, Manchester, United Kingdom.}%
}

\begin{document}

\maketitle
\thispagestyle{empty}
\pagestyle{empty}

\begin{abstract}     
In this paper we consider the problem of allowing a humanoid robot that is subject to a persistent disturbance, in the form of a payload-carrying task, to follow given planned footsteps. To solve this problem, we combine an online nonlinear centroidal Model Predictive Controller - MPC  with a contact stable force parametrization. The cost function of the MPC is augmented with terms handling the disturbance and regularizing the parameter. The performance of the resulting controller is validated  both in simulations and on the humanoid robot iCub. Finally, the effect of using the parametrization on the computational time of the controller is briefly studied.
\end{abstract}

\smallskip 

\section{Introduction}

\begin{figure}[t]
\centering
\includegraphics[height=\columnwidth]{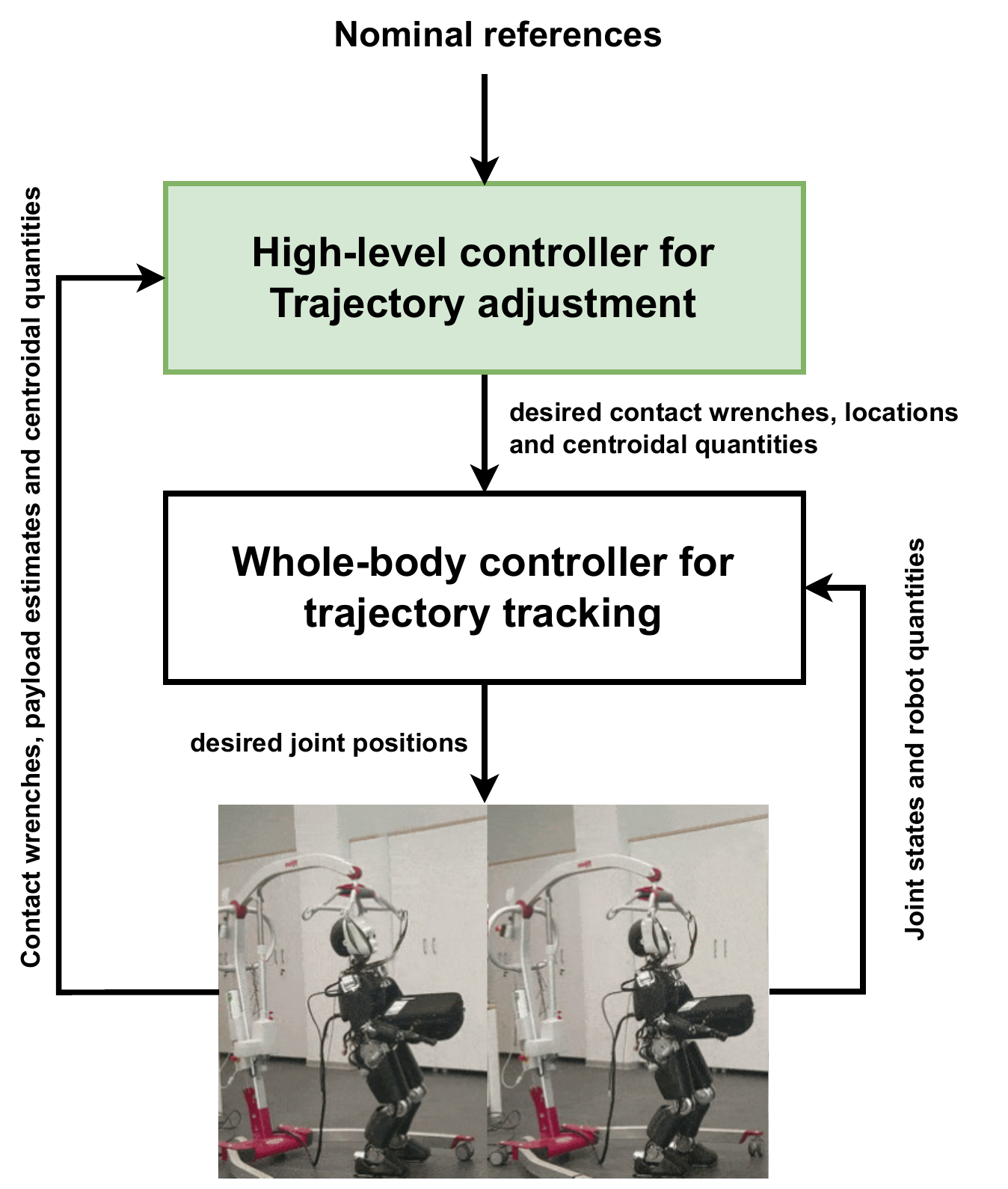}
\caption{The controller highlighted in a typical multi-layer bipedal locomotion control architecture.}
\label{fig:architecture}
\end{figure}

Of the various humanoid robots locomotion control architectures, a modular model-based design emphasizing the separation between high-level trajectory generation and adjustment block(s) and a whole-body trajectory tracking block is becoming a mainstay \cite{Koolen}. The high-level control layer typically utilizes \enquote{template} models to reason about the center of mass and feet trajectories \cite{GiulioICRA}, while the whole-body control layer uses the robot full model to track the adapted trajectories (see Figure \ref{fig:architecture}).  This paper focuses on designing a high-level trajectory adjustment controller leveraging a template model to allow for humanoid robots locomotion under the action of persistent disturbances. 

When using  \enquote{template}  models for high-level objectives, the Linear Inverted Pendulum (and its variations) \cite{LIPM} is typically used in an optimal control formulation \cite{BenchmarkingGiulio}. However, while this approach proved very successful attested by the large body of literature on the subject \cite{englsberger,griffin_ihmc, benchmarking_laas}, it limits the engineer both in terms of how dynamic the walking pattern achieved is (specially for position controlled robots) as well as the different tasks possible (e.g. running and jumping). In this context reduced models (of which Orin's and Goswami's centroidal momentum dynamics model \cite{Orin} is perhaps the most relevant to our work) are more suited to different balancing and locomotion tasks. 

Moreover, as humanoid robots are increasingly expected to carry out collaborative tasks in a work environment \cite{lorenzoICRA}, research on \textit{robust locomotion} for humanoid robots is gaining more interest \cite{wieber, Griffin, grizzle}. The collaborative nature of the work entails studying not only impulsive disturbances effects on the humanoid balancing and locomotion (e.g., a push on the humanoid robot \cite{push_recovery, Jeong}), but also \textit{persistent} disturbances (e.g., a payload being carried by the robot \cite{ Harada_1,Harada_2, Kheddar, collaborative_walking}).  

Several works in the literature deal with persistent disturbances in quadruped (e.g. \cite{hutter}) and humanoids locomotion (e.g. \cite{diehl, Villa, smaldoneHumanoids}). In particular, they can be classified in two broad categories; the first approach deals with the disturbance via robust control machinery. An example of this is the use of robust tube-based MPC as in the work of Villa and Wieber \cite{Villa}.

The second (and more relevant to this work) approach relies on leveraging some knowledge about the persistent disturbance in the controller design. This knowledge may be obtained via the use of disturbance observers and/or estimators as in the works of Smaldone et. al \cite{smaldoneHumanoids}, Kaneko et. al \cite{Kaneko} and the references therein.

In this direction, this paper contributes towards allowing a humanoid robot to carry a payload while walking, leveraging knowledge of the payload effects on the centroidal dynamics quantities. More in details, building upon \cite{GiulioICRA}, we propose an online non-linear MPC \textit{payload-aware} high-level controller. In addition, we simplify the online non-linear optimization problem by eliminating the so-called contact-stability constraints \cite{Featherstone} on the feet. This simplification is achieved thanks to a recently introduced contact-stable force parametrization \cite{NavaTRO}. Consequently, the optimization variable is changed and the number of constraints is reduced transferring the complexity to the cost function by adding terms specifically handling the payload effect. The non-linear MPC problem is solved online and validated on the example of a floating mass with two legs as well as on the humanoid robot iCub \cite{icub}. Furthermore, the computational effort required is briefly investigated specifically in terms of the effect of removing constraints from the original optimization problem through the use of the contact-stable force parametrization. 

The manuscript is organized as follow; Section \ref{sec:prel} introduces the notations then summarily recalls the necessary machinery used throughout this work and states the problem treated in an explicit manner. Section \ref{Sec:three} introduces the proposed controller solving our problem together with some necessary comments and remarks. Section \ref{sec:validation} focuses on the validation of the proposed controller both in terms of simulations and experimental results. Finally, some concluding remarks in Section \ref{sec:conclusions} end the manuscript.

\section{Background}\label{sec:prel}
In this section, we first introduce the notations used throughout, then report the basic ingredients required for designing our proposed controller together with a statement of the problem addressed.

\subsection{Notations and nomenclature}  For any vector $z \in \mathbb{R}^n$, $\|z\|$ and $z^\top$ define respectively the norm and transpose of $z$. Given a full rank matrix $B \in\mathbb{R}^{n\times m}$ with $n>m$,  $B^\dagger = (B^\top B)^{-1}B^\top$ denotes the Moore-Penrose inverse, while $B^\perp$ its orthogonal complement verifying $B^\perp B = 0$,  and $\text{ker}\{B\}$ denotes its null space. $I_n$ and $0_n$ denote the identity and zero matrices of dimension $n$. For $x \in \mathbb R^3$, $x^\wedge = S(x) : \mathbb R^3 \to \mathfrak{so(3)}$ returns the skew symmetric matrix form of $x$.  
Additionally, the following standard nomenclature will be used throughout the text;
\begin{itemize}
    \item $H^{\ell}, \ H^{\omega}$ will denote the \textit{aggregate} linear and angular momentum of all links referred to the robot center of mass CoM and oriented as the inertial frame.
    \item $P_{CoM}$ is the position of the CoM referred to the inertial frame.
    \item $P_{\mathcal{C}_i}, \ R_{\mathcal{C}_i}, \ v_{\mathcal{C}_i}$ will represent the position of a contact point $i$ referred to the inertial frame, a rotation matrix expressing the orientation of a frame attached to the contact point and the linear contact velocity respectively.
    \item For each foot we have $n_c \in \{1, 2\}$ the total number of possible contacts at each stance phase;
    \item $n_v = 4$ is the number of contact vertices assuming a fixed rectangular contact surface.
    \item $\gamma_i \in \{0, 1 \}$ is a variable capturing whether contact $i$ is active or not.
    \item Finally, $m$ is the robot mass and the gravity acceleration vector is denoted $\ \Vec{g} = \begin{pmatrix}
       0 & 0 & 9.81 & 0 & 0 & 0
    \end{pmatrix}^\top$.
\end{itemize}

\subsection{The centroidal momentum dynamics}

The \textit{reduced} dynamics of the centroidal quantities of a humanoid robot carrying a payload can be modeled in the standard non-linear control affine perturbed form \cite{Isidori};
\begin{align}\label{sys_ct}
    \dot x &= f(x) + G(x)u + p(x)d
\end{align}
where $d$ corresponds to the external wrench(s) caused by the payload, and the drift, control and disturbance vector fields take the forms (omitting the zero block-matrix dimensions for clarity);
\begin{subequations}
\begin{align}
    f(x) &= \begin{pmatrix}
       f_1(x) \\ f_2(x) \\ 0
    \end{pmatrix}, \ G(x) = \begin{pmatrix}
       0 & 0\\ g_{2,1}(x) & 0 \\ 0 & g_{3,2}(x)
    \end{pmatrix}, \\  p(x) &= \begin{pmatrix}
       0 \\ p_2(x) \\  0
    \end{pmatrix}
\end{align}
\end{subequations}
where, following \cite{GiulioICRA}, we set
\begin{subequations}
\begin{align}
    x &= \begin{pmatrix}
       x_1 \\ x_2 \\ x_3
    \end{pmatrix} = \begin{pmatrix}
       P_{CoM}\\ H \\ P_{\mathcal{C}}
    \end{pmatrix}, \  H = \begin{pmatrix}
       H^{\ell} \\ H^\omega
    \end{pmatrix} \\
    u &= \begin{pmatrix}
       u_1 \\ u_2
    \end{pmatrix} = \begin{pmatrix}
       f \\ v_{\mathcal{C}}
    \end{pmatrix}
\end{align}
\end{subequations}
for $f, v_{\mathcal{C}}$ vectors collecting external wrenches at the feet and contact velocities, and for each contact, capturing the contact status by letting
\begin{align}
    \dot{P_{\mathcal C_i}} &= (1-\gamma_i)v_{\mathcal C_i}
\end{align}
Performing the above assignment, we have explicitly;
\begin{align*}
    f_1(x) &= \frac{1}{m}Bx_2, \quad B = \begin{pmatrix}
       I_3 & 0_3
    \end{pmatrix}\\
    f_2(x) &= -m\Vec{g}\\
    g_{2,1}(x) &= A_f \Gamma, \quad g_{3,2}(x) = I-\Gamma \\
    p_{2}(x) &= A_d
\end{align*}
where $\Gamma$ is a diagonal matrix of appropriate dimensions with diagonal blocks corresponding to $\gamma_i$ contacts, and the matrices
\begin{align*}
    A_f &= \begin{pmatrix}
   A_{f_1} \hdots A_{f_{n_c \times n_v}} \end{pmatrix}\\
   A_d  &= \begin{pmatrix}
      A_{d_l} & A_{d_r}
   \end{pmatrix}\\
   A_{f_i} &= \begin{pmatrix}
      I_3 & 0_3 \\ (P_{\mathcal{C}_i} - P_{CoM})^\wedge & I_3
   \end{pmatrix} \\
   A_{d_j} &= \begin{pmatrix}
      I_3 & 0_3 \\ (P_{d_j} - P_{CoM})^\wedge & I_3
   \end{pmatrix}
\end{align*}
with $P_{d_j}, j \in \{ l, r \}$ the contact point of the payload, assumed fixed on the left and right hands respectively. 
The dynamics (\ref{sys_ct}) will be used for controller design in Section \ref{Sec:three}.

\subsection{The contact-stable force parametrization}
For a given contact \textit{wrench};
\begin{align*}
    w &= \begin{pmatrix}
       F_x & F_y & F_z & M_x & M_y & M_z
    \end{pmatrix}^\top
\end{align*}
where $F_i, M_i$ are the corresponding force and moment in the $i$ direction respectively, maintaining stable planar unilateral contact is tantamount to enforcing the following constraints;
\begin{subequations}\label{stable_contact_constraints}
\begin{align}
    F_z &> f_z^{\text{min}} \geq 0\\
    \sqrt{F_x^2 + F_y^2} &< \mu_c F_z\\
    y_c^{\text{min}} &< \frac{M_x}{F_z} < y_c^{\text{max}}\\
    x_c^{\text{min}} &<- \frac{M_y}{F_z} < x_c^{\text{max}}\\
    \left| \frac{M_z}{F_z} \right| &< \mu_z
\end{align}
\end{subequations}
in which $x_c^{\text{min}}, x_c^{\text{max}}, y_c^{\text{min}}, y_c^{\text{max}}$ are the dimensions of the herein assumed rectangular contact surface, and $\mu_c, \mu_z$ are the static and torsional friction coefficients respectively. Constraints (\ref{stable_contact_constraints}) above define a constraints set $\mathcal{K}$, and can be equivalently written as $w \in \mathcal{K}$.

In \cite{NavaTRO}, a stable contact wrench parametrization 
$$ w = \phi(\xi): \mathbb{R}^6 \mapsto \tilde{\mathcal{K}} \subset \mathcal{K}$$ for some $\xi \in \mathbb{R}^6$ was proposed ensuring the satisfaction of (\ref{stable_contact_constraints}), where
\begin{align}\label{parametrization}
    \phi(\xi) = \begin{pmatrix} \mu_c \frac{\tanh \xi_1 (e^{\xi_3}+ f_z^{\text{min}})}{\sqrt{1 + \tanh^2 \xi_2}}\\  \mu_c \frac{\tanh \xi_2 (e^{\xi_3}+ f_z^{\text{min}})}{\sqrt{1 + \tanh^2 \xi_1}} \\ e^{\xi_3} + f_z^{\text{min}} \\ (\delta_{y}\tanh\xi_4 + \delta_{y_0})(e^{\xi_3} + f_z^{\text{min}}) \\ (\delta_{x}\tanh\xi_5 + \delta_{x_0})(e^{\xi_3} + f_z^{\text{min}}) \\ \mu_z \tanh (\xi_6)(e^{\xi_3} + f_z^{\text{min}})   \end{pmatrix}
\end{align}
and
\begin{align*}
    \delta_x &= \frac{x_{c}^{\text{max}}- x_{c}^{\text{min}}}{2}, \quad \delta_{x_0} = -\frac{x_{c}^{\text{max}} + x_{c}^{\text{min}}}{2}\\
\delta_y &= \frac{y_{c}^{\text{max}}- y_{c}^{\text{min}}}{2}, \quad \delta_{y_0} = \frac{y_{c}^{\text{max}} + y_{c}^{\text{min}}}{2}
\end{align*}
\begin{remark}\label{parametrization_properties}
given the contact constraints (\ref{stable_contact_constraints}), the parametrization (\ref{parametrization}) represents an almost global cover of the corresponding feasible contact wrenches. The constraints (\ref{stable_contact_constraints}) covers our case, assuming no flight phase during walking, and a co-planar contact at the footsteps impacts, and consequently this parametrization suffices. The interested reader is referred to \cite{NavaTRO} for the technical details. 
\end{remark}
\subsection{Problem statement}

The problem we are concerned with can now be stated;

\begin{problem}\label{problem_statement}
Given the dynamics (\ref{sys_ct}), and a set of \textit{nominal} footstep locations and corresponding CoM (and momentum) references, it is required to design a feedback control law such that the following is achieved;
\begin{enumerate}
    \item \textbf{Nominal references tracking}: a bounded tracking error on both the centroidal quantities nominal references and contact locations is maintained $\forall t$.
    
    \item \textbf{Disturbance attenuation}: the effect of the payload on the tracking performances is \textit{minimized} \footnote{such that, barring mechanical limitations of the hardware, the robot is able to carry and walk with the payload without falling over}.
\end{enumerate}
\end{problem}

It is imperative to note that the tracking requirement is relaxed compared to the exact tracking problem version. This is also motivated by the need to allow the feedback to provide a footstep adjustment capability. It is also clear that we are not asking, explicitly, for the feedback to respect some actuator limits, since we implicitly assume that this is taken care of by the whole-body control layer as  in Figure \ref{fig:architecture}. In the next section, we will detail our solution approach to this problem.

\section{Non-linear centroidal MPC with contact-stable force parametrization}\label{Sec:three}

Starting from the multi-layer architecture in Figure \ref{fig:architecture}, we propose to modify the high-level trajectory adjustment layer, utilizing a receding horizon approach \cite{grune}. Indeed, based on the recently proposed online non-linear centroidal MPC controller \cite{GiulioICRA}, we propose to modify the prediction model, the cost function and constraints in order to solve the problem.

\subsection{The cost function}
The cost function reflects the objectives detailed in the problem statement, and to this end, the following tasks are considered in the cost function.

\subsubsection{The nominal references tracking task} this can be further decomposed, rewriting
\begin{align}
    T_1 &= T_h + T_{P_{\mathcal C}}
\end{align}
where the centroidal dynamics task $T_h$ reads;
\begin{align}\label{T_h}
    T_h = \frac{1}{2}\| L x_2\|_{Q_h}^2 + \frac{1}{2}\| x_1 - x_1^n\|_{Q_c}^2 
\end{align}
with $L = \begin{pmatrix}
   0_3 & I_3
\end{pmatrix}$ and  $x_1^n$ a reference trajectory for the CoM position, while  $Q_h , Q_c \geq 0$  positive definite penalty matrices of suitable dimension.

As for the footstep location tracking task $T_{P_{\mathcal C}}$, given a nominal reference $x_3^n$ (e.g., coming from an external planner block), we incorporate the following penalty; 
 \begin{align}\label{T_pc}
     T_{P_{\mathcal C}} &= \frac{1}{2}\| x_3 - x_3^n \|_{Q_{P_C}}^2  
 \end{align}
 in which $Q_{P_C} > 0$ is a positive definite penalizing weight.

\subsubsection{The Payload attenuation task} this can be further decomposed into;
\begin{align}
    T_2 &= T_d + T_\xi
\end{align}
where the payload attenuation (and gravity compensation) term $T_d$ takes the form
\begin{align}\label{T_d}
    T_d &= \frac{1}{2} \| u_1  -   (A_f)^{\dagger}(-A_d d) - \frac{1}{n_c \times n_v}m\Vec{g} \otimes I_{n_c \times n_v}\|_{Q_d}^2
\end{align}
 rewriting $ u_1 = \phi(\xi)$ in terms of the parameter (our optimization variable) and $Q_d > 0$ a positive definite penalizing matrix of suitable dimension.  In addition, we add a regularizing task $T_\xi$ on the parameter value of the form;
\begin{align}\label{T_xi}
    T_\xi &=  \frac{1}{2}\| \xi  \|_{Q_\xi}^2
\end{align}
where $Q_\xi > 0$ is a positive definite penalizing matrix. 

\begin{remark}\label{xi_regularization}
the regularizing task on the parameter $\xi$ magnitude stems from the fact that
\begin{align*}
    \lim_{\xi \to 0 }\phi(\xi) = \begin{pmatrix}
       0 & 0 & 1+f_z^{\text{min}} & 0 & 0 & 0
    \end{pmatrix}^\top
\end{align*}
with the contact force application being in the interior of the support surface. In other words, as the parameter decreases, the corresponding contact force has no components parallel to the contact surface and are always in the interior of the feasible set. 
\end{remark}
\subsection{The constraints}

The parametrization (\ref{parametrization}) allows us to eliminate the contact-stability constraints (\ref{stable_contact_constraints}) from our optimization problem. In this sense, and differently from \cite{GiulioICRA}, we only apply inequality constraints corresponding to restricting the maximum allowable footstep location error. Namely we set;
\begin{align}
    -lb \leq \| R_{\mathcal{C}_i} (x_3^i - x_3^{n_i}) \| \leq ub
\end{align}
where, as in (\ref{T_pc}), $x_3^{n_i}$ is a nominal reference for the $i$ contact vertex and $lb, \  ub$ are lower and upper bounds for the tracking error respectively.  
\subsection{The general MPC problem}
With all the ingredients in place, we can state the general MPC problem to be solved online for $i = 1, \hdots, n_p$; 
\begin{subequations}\label{mpc_problem}
\begin{align}
V^\star &= \text{min} \ \ \sum_{k = 0}^{n_p}  T_1 + T_2 \\
 \text{s.t} \quad & \notag\\
 x(k+i) &= F(x(k+i-1), u(k+i-1)) \label{prediction_model} \\
    -lb &\leq \| R_{\mathcal{C}_i} (x_3(k+i) - x_3^n(k+i) \| \leq ub
\end{align}
\end{subequations}
with $n_p$ being the so-called prediction horizon and where the prediction model (\ref{prediction_model}) is obtained by discretizing the dynamics (\ref{sys_ct}) using any numerical integrator \cite{Franklin}. For our validation purposes, we use a simple Euler's discretization scheme. The controller outputs are generated using the Receding Horizon Principle, adopting a fixed prediction window with a length equal to $n_p$ samples. Specifically, solving (\ref{mpc_problem}) for the minimizer $(\xi^\star, \ v_{\mathcal{C}}^\star)$ then the feedback solution to Problem \ref{problem_statement} reads;
\begin{align}\label{feedback}
    u^\star &= \begin{pmatrix}
       \phi(\xi^\star)^\top & v_{\mathcal{C}}^{\star^\top}
    \end{pmatrix}^\top
\end{align}

\begin{remark}\label{terminal_ingredients}
notice that no terminal ingredients (e.g. terminal penalty and constraints set \cite{grune}) are added in this MPC problem to ensure closed loop recursive feasibility and stability. This is beyond the scope of this article.
\end{remark}

\subsection{The payload estimation}
At any given time instant, the payload external wrench $d(t)$ and its contact location $P_{d_i}(t)$ is needed for $n_p$ steps ahead in the prediction model to solve the optimization problem (\ref{mpc_problem}). This can be satisfied by having a simple dynamics modelling the evolution of both $d(t)$ and $P_{d_i}(t)$ and integrating over time. In control-theoretic terms, an \textit{exo-system} generating the references and disturbance trajectories can be used \cite{byrnes}. Instead, we opt to simplify this requirement in the following manner;

\begin{figure*}[!h]
	\centering
	\subfloat[\centering Parametrized MPC with payload attenuation task]{{\includegraphics[width=0.5\textwidth]{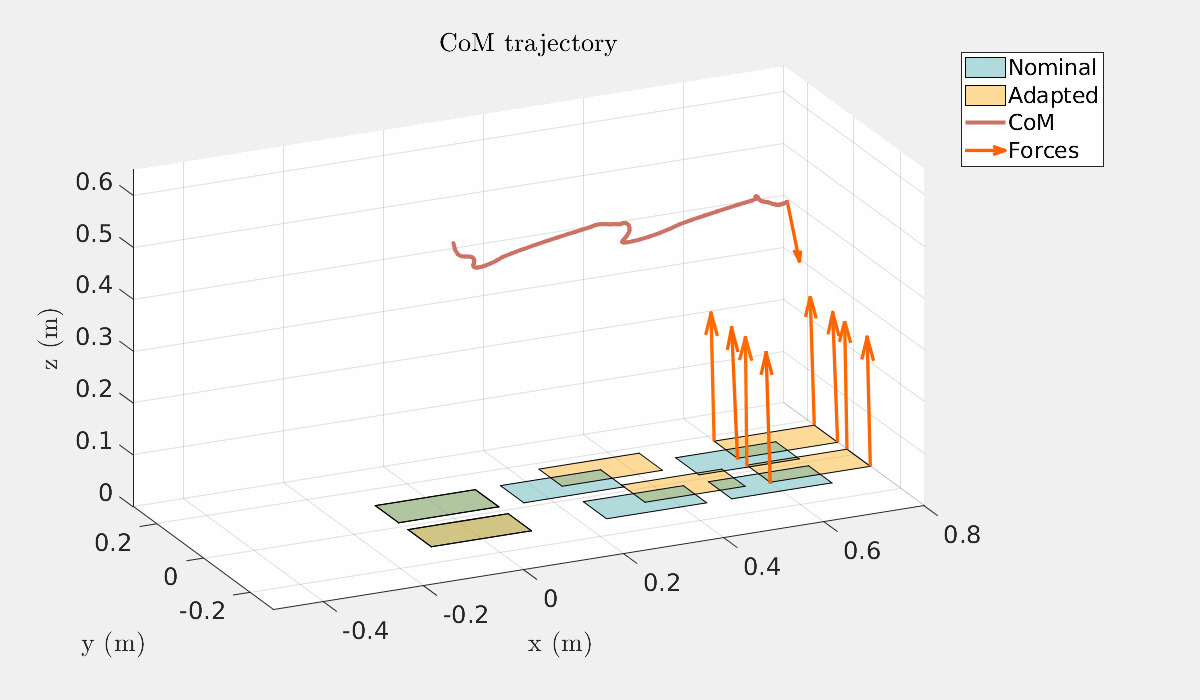} }}%
	\subfloat[\centering Non parametrized MPC without payload attenuation task]{{\includegraphics[width=0.5\textwidth]{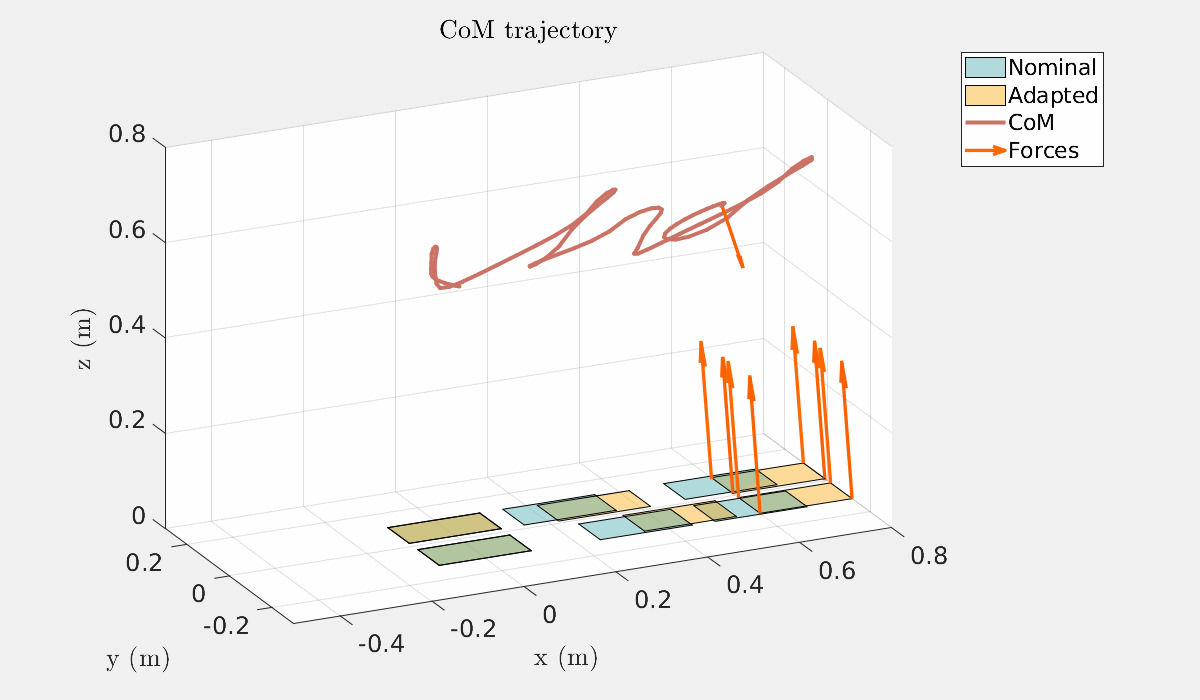} }}%
	\caption{Comparison of obtained CoM trajectories and adapted contact locations of the proposed parametrized MPC with payload attenuation (a), and a non-linear centroidal MPC without payload attenuation (b). The payload external force acting on the CoM is also depicted with an orange arrow, together with the feet contact forces.}%
	\label{fig:academic_example}%
\end{figure*}

\begin{assumption}\label{assumption_payload}
For $t = k\delta$, $\delta$ being the sampling period used to integrate the prediction model, and given an \textit{estimate} of the payload external wrench $\hat{d}(k)$ the following holds;
\begin{align*}
    d(k+i) = \hat{d}(k), \quad i = 1 \dots n_p 
\end{align*}
Moreover, we assume the payload contact location w.r.t. CoM to be given by $P_{d_r}(k+i) = \bar{P}_{d_r}(k), \ P_{d_l}(k+i) = \bar{P}_{d_l}(k)$ for $i = 1 \dots n_p$ where $\bar{P}_{d_r}(k), \ \bar{P}_{d_l}(k)$ are the right and left hands positions at time $k$ respectively. 
\end{assumption}

It is worth stressing that this is a valid assumption since the payload disturbance is updated in the next iteration of the MPC thanks to the receding horizon implementation.

When testing on the robot,  both $\hat{d}(k)$ and $\bar{P}_{d_i}(k)$ are needed. For the payload contact location, the robot kinematics model is enough to obtain, for each time step, the values of $\bar{P}_{d_i}$. As for the payload external wrench, we use an estimation algorithm that relies on propagating the measurements obtained by a given set of force-torque sensors through the kinematic chain, writing and aggregating the Newton’s-Euler’s equations for each rigid link in the sub-chain and solving for the external wrench via a least-square norm-minimization-like technique (for more details the reader is referred to \cite{wbd_ref} and the references therein). 

To conclude this section, some remarks are in order:
\begin{itemize}
    \item  the task (\ref{T_d}) results from applying direct disturbance compensation techniques. When in double support, there may exists infinitely many values for the parameter $\xi$ satisfying the requirement. A projection on the null space to enforce a given selection of the minimizer may be be used \cite{NavaTRO}. The current set-up however does not enforce a strict prioritization of (\ref{T_d}). 
    \item neglecting the disturbance compensation (\ref{T_d}) and parameter regularization (\ref{T_xi}) tasks, the MPC problem (\ref{mpc_problem}) is not necessarily equivalent to that in Section III-D of \cite{GiulioICRA}. Consequently, as explained in Remark \ref{parametrization_properties}, there exist feasible feet contact wrenches solving the original problem, for which no optimum parameter solving (\ref{mpc_problem}) can be obtained. However, for our purposes, the paramtrization covers almost all ($90\%$ of) the feasible forces space \cite{NavaTRO}.
\end{itemize}

\section{Results and validation}\label{sec:validation}

In this section, we validate the proposed controller on several scenarios and comment on the obtained results. In all cases, we use Casadi$^\copyright$ \cite{casadi}, IPOPT \cite{ipopt} and the linear solver HSL MA97$^\copyright$ \cite{HSL} to solve the optimization problem (\ref{mpc_problem}). In addition, table \ref{tab:params} list the various parameters and weight penalties that are fixed throughout the validation case studies. The proposed controller runs on a  10th generation Intel$^\copyright$ Core i7-10750H laptop equipped with Ubuntu Linux 20.04. For bench-marking reasons, and whenever mentioned, we compare the modified payload-aware controller against the MPC controller presented in \cite{GiulioICRA}. This is done using parameters and penalizing weights that were originally tuned for the controller in \cite{GiulioICRA}.

\begin{table}[t]
\begin{tabular} {| c | c | c | c | }
 \hline
$n_p$ [time steps] & 10  & MPC rate $[s]$ & 0.2\\ 
\hline
$\mu_c$ $[N/A]$ & 0.33 & $f_z^{\text{min}}$  & 0.01  \\  
\hline
$Q_h$ & $10^2 \times I_3 $ & $Q_{c}$  & $\text{diag}\{ 1, \ 1, \ 10^3 \}$  \\  
				\hline
$Q_{P_c}$ &  $200 \times I_3 $ & $Q_d$ &  $10^2 \times I_3 $  \\   
				\hline
				$Q_\xi$ & $10 \times I$ &  &   \\ 
\hline
\end{tabular}
\caption{\label{tab:params} parameters and penalizing weights}
\end{table}
\subsection{Reduced models simulations}
We first validate the proposed approach in the case of a floating mass equipped with two legs. In this case we set  $m = \SI{1}{\kilo \gram}, \ n_c \in \{ 1, 2\}, n_v = 4$ and the CoM height being $\bar{c_z} = \SI{0.53}{\meter}$. The system is subject to a $\SI{1.5}{\kilo \gram}$ payload at an assumed fixed payload contact location being $P_{d_i} = \begin{pmatrix} 0.25 & \pm 0.1 & -\frac{\bar{c_z}}{4} \end{pmatrix}^\top$ for the left and right payload contact location w.r.t. to the CoM respectively. Figure \ref{fig:academic_example} depicts the generated CoM trajectories and foot-step locations and contact forces in this case. 

It is clear from the simulation figures that including an the payload attenuation task in the MPC problem results in more reasonable CoM trajectory, i.e. with less variation in the CoM height for the same tuning parameters. More interestingly, while the feet contact forces are comparable in both cases, the proposed parametrized centroidal MPC compensates for the payload with larger steps adaptation.

\subsection{Validation on the iCub3 robot}

To test the proposed controller performances on a humanoid robot, we use the iCub3 \cite{icub3_avatar}, a $\SI{1.25}{\meter}$ tall, and weighs $\SI{52}{\kilo \gram}$. The tests were carried first in simulation using Gazebo \cite{gazebo} and then replicated on the real hardware (modulo sensor/actuator noise and wrist mechanical construction limits). Indeed, in this case, the payload contact locations are obtained online at each time step using the robot kinematics assuming a fixed contact location at the hand reference frame. Additionally, and as is done in \cite{GiulioICRA}, the whole-body controller is a Quadratic Program - QP over the robot generalized velocity and the joint velocities. The QP program is solved using an off the shelf solver. 

\begin{figure*}[h!]
	\centering
	{{\includegraphics[width=0.85\textwidth]{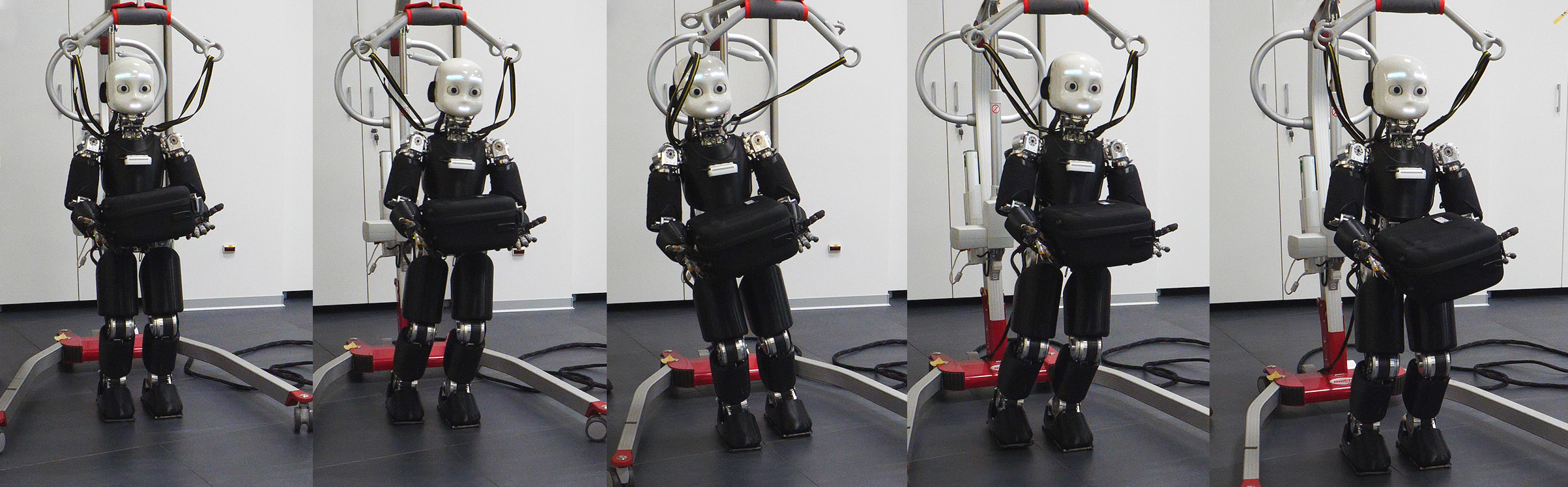} }}%
    \begin{multicols}{2}
        \includegraphics[width=\linewidth]{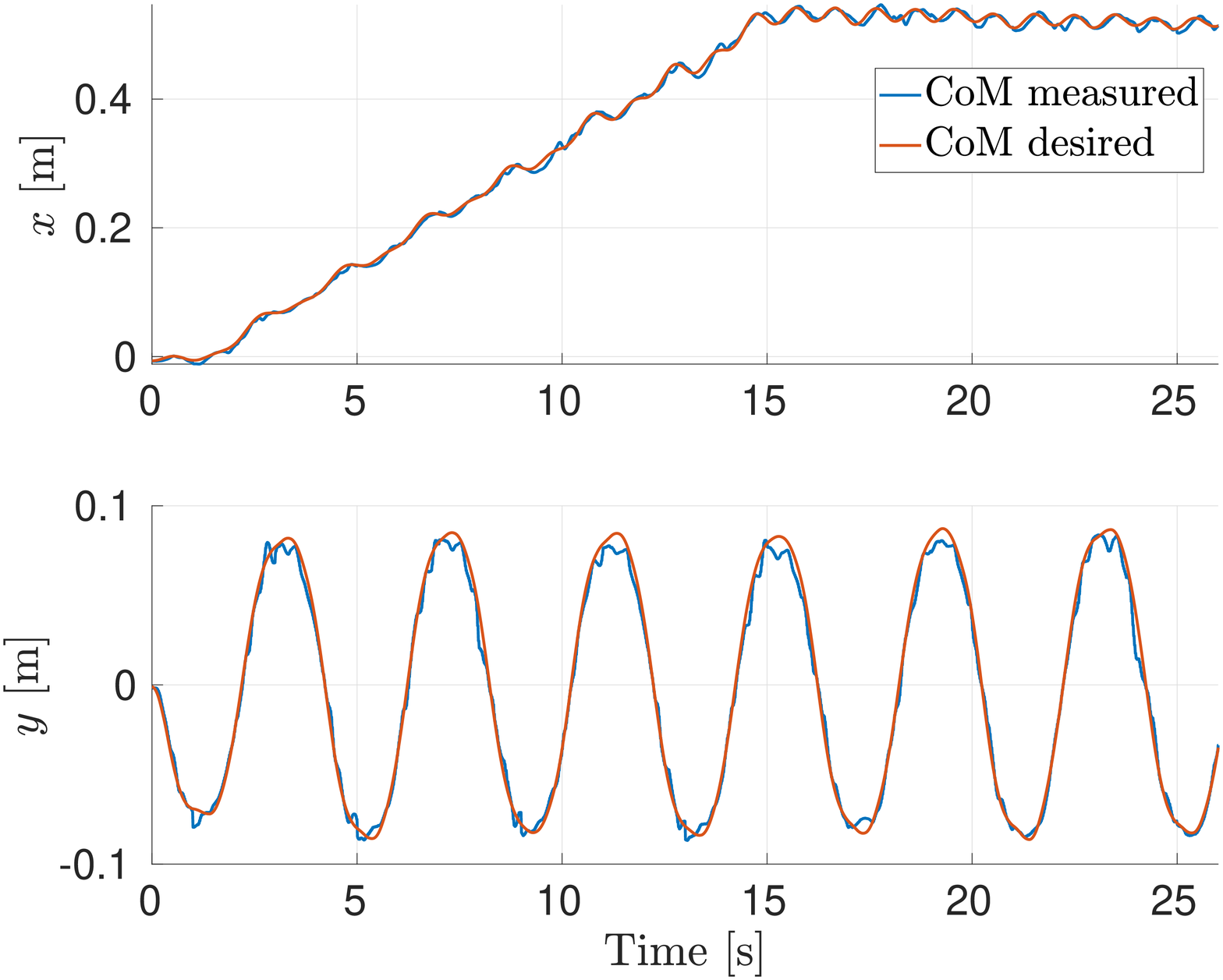}
        \includegraphics[width=\linewidth]{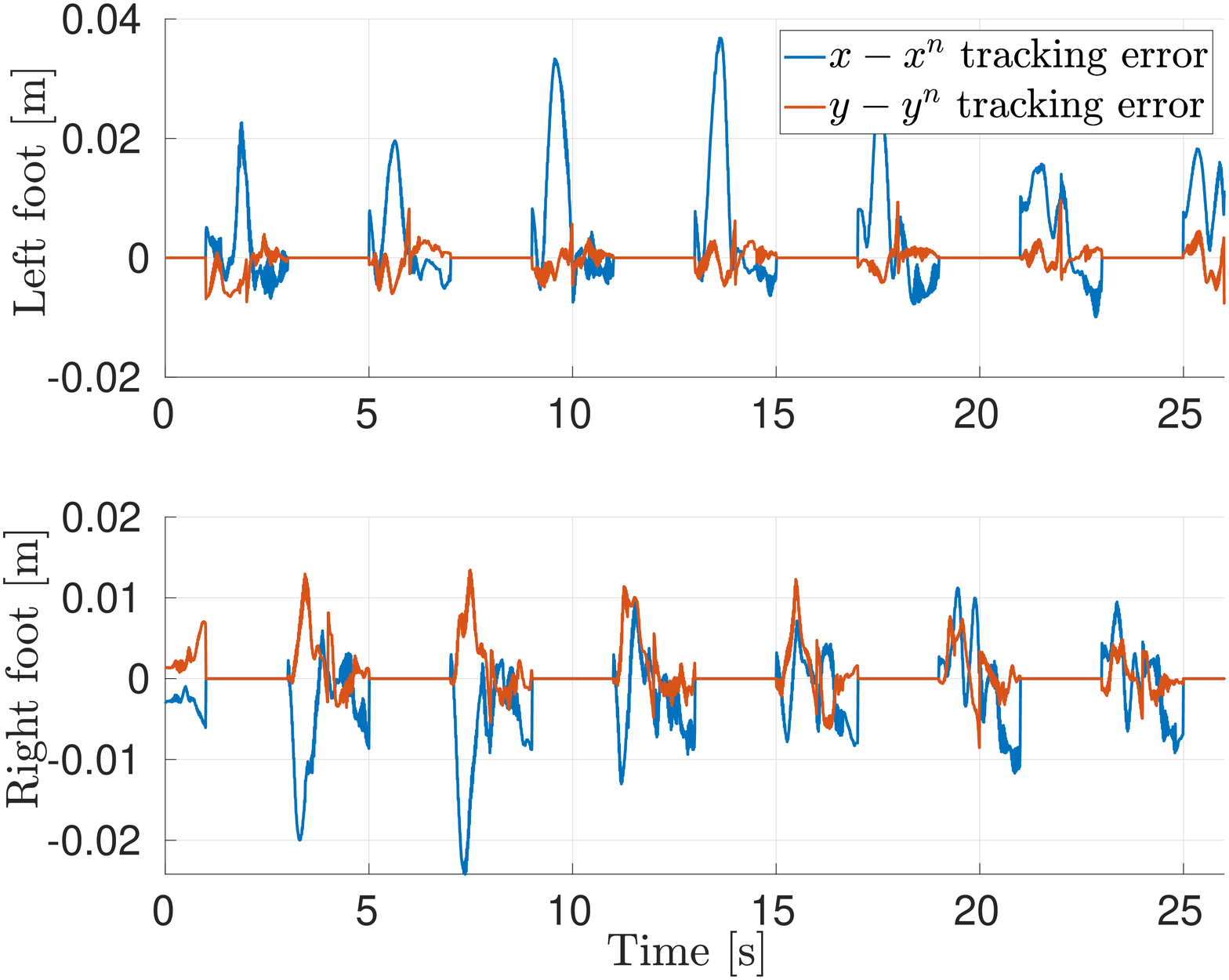}
    \end{multicols}         
	\caption{Top: screenshots of the robot walking while carrying the payload. Bottom left: the CoM tracking in closed loop. Bottom right: the tracking errors of the generated (adapted) feet contact locations w.r.t the nominal ones.}%
	\label{fig:tracking_performances}%
\end{figure*}

In this scenario, we give the robot planned nominal feet steps and corresponding trajectories. In addition, the robot is carrying a box that contains a small weight of $\SI{2}{\kilo \gram}$ (due to the iCub3 wrist construction limitations \footnote{In simulations, as explained in the accompanying video, we are able to attach a mass of $m = \SI{10}{\kilo \gram}$. Related code \url{https://github.com/ami-iit/paper_elobaid_2023_icra_walking_with_payloads.git}}). Figure  \ref{fig:tracking_performances} illustrates the obtained performances in terms of tracking of the CoM and the feet contact locations adaptation. As expected, the CoM tracking error is about $\SI{0.01}{\meter} - \SI{0.04}{\meter}$ with persistent adaptations on the feet tracking to compensate for the payload (i.e with a peak of about $\SI{0.021}{\meter}$). The robot manages to conclude the task successfully.

\subsection{Comments on computational aspects}

One interesting question concerning the simplification of the MPC problem by using the parametrization (\ref{parametrization}) to reduce the number of constraints would be; \textit{what are the gains, if any, in terms of computation time and number of iterations?}

\begin{figure}[!h]
    \centering
    \includegraphics[width=0.45\textwidth]{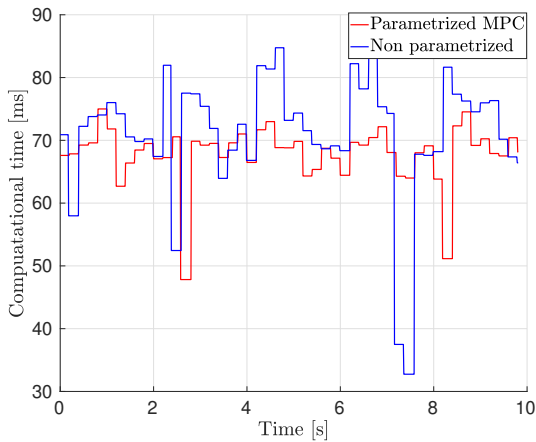}
    \caption{Computational time comparison}
    \label{fig:compuatations}
\end{figure}

To fairly judge this aspect, we removed the corresponding payload attenuation (\ref{T_d}) and  parameter regularization (\ref{T_xi}) tasks. Those instead were replaced with a force regularization term rendering the feet contact forces as similar as possible. In this sense, we are able to compare the parametrized centroidal MPC against the original version developed in \cite{GiulioICRA}.  Figure \ref{fig:compuatations} illustrates the computational time in both problems in one run where the objective is to allow the robot to walk following some nominal planned footstep trajectory. This direct comparison shows that while in a lot of time steps, the original centroidal MPC out-performs the proposed controller in terms of computation time required, however on average, the proposed controller gives better results ($\SI{0.068}{\second}$ compared to $\SI{0.073}{\second}$). It is also clear that the performance gain is not by much. In essence, removal of constraints in a non-linear optimization problems does not necessarily lead to \textit{significant}, if any, computational gains.
\section{Conclusions and perspectives}\label{sec:conclusions}
This work presented an online non-linear centroidal MPC controller in combination with an external payloads estimates and a stable-force parametrization to allow a humanoid robot to carry objects while walking. The \enquote{robustness} of the proposed controller stems from a specifically designed term in the cost function that rejects the effect of the disturbance, as well as including the payload effects on the centroidal quantities dynamics in the prediction model.  The controller is validated both in simulations and on a newer version of the humanoid robot iCub. Being a simplified MPC controller thanks to the force parametrization, some computational gains are obtained, albeit being rather small.

As a future direction of research, we are investigating additional avenues of lowering the computational time required. In addition, we are interested in closing the loop on the contact forces to achieve better contact forces tracking. 

\section*{Acknowledgment}
This work was partially supported by the Italian National Institute for Insurance against Accidents at Work (INAIL) ergoCub Project, and Honda Research and Development Japan, through a joint-lab research initiative.

\balance
\bibliographystyle{IEEEtran}      
\bibliography{biblio}                  

\end{document}